\pdfoutput=1 
\documentclass[11pt,a4paper]{article}
\usepackage[hyperref]{style/naaclhlt2018}

\usepackage{times}
\usepackage{amsmath}
\usepackage{latexsym}
\usepackage{csquotes}
\usepackage{fixltx2e}
\usepackage{graphicx}
\usepackage{tikz}
\usepackage{amssymb}
\usepackage{color}
\usepackage[normalem]{ulem} 
\usepackage[vlined, ruled, boxed, linesnumbered]{algorithm2e}
\usepackage{url}
\usepackage{multirow}
\usepackage{booktabs}

\aclfinalcopy 

\title{Multi-Task Learning for Argumentation Mining in Low-Resource Settings}

\author{Claudia Schulz, Steffen Eger, Johannes Daxenberger, Tobias Kahse, Iryna Gurevych \\
  Ubiquitous Knowledge Processing Lab  (UKP-TUDA)\\ Department of Computer Science\\ Technische Universit\"{a}t Darmstadt \\
  {\tt www.ukp.tu-darmstadt.de} }

\date{}

\begin{document}

\maketitle
\begin{abstract}
We investigate whether and where multi-task learning (MTL)
can improve performance on 
NLP problems related to
argumentation mining (AM),
in particular
argument component identification.
Our results 
show that MTL 
performs particularly well (and better than single-task learning) when little training data is available for the main task, a common scenario in AM. 
Our findings challenge previous assumptions that conceptualizations across AM datasets are divergent and that MTL is difficult for semantic or higher-level tasks.
\end{abstract}

\section{Introduction}\label{sec:introduction}
Computational argumentation mining (AM) deals with the automatic identification of argumentative structures within natural language.
This can be beneficial in many applications such as summarizing arguments in texts to improve comprehensibility for end-users, or information retrieval and extraction \cite{Persing2016}. 
A common task is to segment a text into argumentative and non-argumentative components and identify the type of argumentative components.
As an illustration, consider the (simplified) example from \newcite{Eger2017}: 
``Since 
[it killed many marine lives]\textsubscript{Premise}
[tourism has threatened nature]\textsubscript{Claim}.''
Here, the non-argumentative token ``Since'' is followed by two argumentative components:
a premise that supports
a claim.

Argumentation is highly subjective and conceptualized in different ways \cite{Peldszus:2013,stein:2017i}.
On the one hand, this implies that creating reliable ground-truth datasets for AM is costly, as it requires trained annotators.
However, even trained annotators have problems identifying and classifying arguments reliably in texts \cite{Habernal2017a}. 
To tackle AM in a new domain or develop new AM tasks, it may thus not be possible to create large datasets as required by most state-of-the-art machine learning approaches.
On the other hand, the different conceptualizations of argumentation resulted in AM corpora with different argument component types, with very little conceptual overlap between some of these corpora \cite{Daxenberger2017}.
This distinguishes AM from more established NLP tasks like discourse parsing \cite{Braud2016} and makes it particularly challenging.
Therefore, a natural question
is how to handle new AM datasets in a new domain and with sparse data.

Here, we investigate how existing AM datasets from different domains and with different conceptualizations of arguments can be leveraged to tackle these challenges. More precisely, we study whether conceptually diverse AM datasets from different domains can help
deal with new AM datasets when data is limited. 
A promising direction to incorporate existing datasets as ``auxiliary knowledge'' is by means of \emph{multi-task learning} (MTL), a paradigm that dates back
to the
1990s \cite{Caruana1993,Caruana1996}, but has only recently gained large attention
\cite{Collobert2011,Sogaard2016,
Hashimoto2017}. 
The idea behind MTL is to learn several tasks jointly, similarly to human learning, so that tasks serve as mutual sources of ``inductive bias'' for one another. 
MTL has been reported particularly beneficial when tasks
exhibit ``natural hierarchies'' \cite{Sogaard2016} or when the amount of training data for the main task is
sparse \cite{Benton2017,Augenstein2017}, where the auxiliary tasks may act as regularizers to prevent overfitting \cite{Ruder2017}. 
The latter is precisely the scenario most relevant to us.

In this paper, we (1)
investigate to which degree training a system to solve several conceptually different AM tasks jointly improves performance over learning 
in isolation,
(2) compare performance gains across different dataset sizes, and
(3) do so across various domains.
Our findings show that MTL 
is helpful for 
AM---particularly in data  sparsity settings---when treating other AM tasks as auxiliary.\footnote{The code and data used for our experiments are available from \url{https://github.com/UKPLab/naacl18-multitask_argument_mining}.}

\section{Related Work}\label{sec:related}
\textbf{AM} is a relatively new field in {NLP}.
Hence, a lot of related work revolves around creating new corpora. We use six such corpora, described in more detail in Section~\ref{sec:experiments}.
On the modeling side, \newcite{Stab2017} and \citet{Persing2016} 
rely on 
pipeline approaches for AM, 
combining parts of the pipeline using integer linear programming (ILP).
\newcite{Eger2017} propose 
neural end-to-end models for {AM}. 
While \newcite{Daxenberger2017}
show that there is little consensus on the conceptualization of a claim across {AM} corpora, 
\newcite{stein:2016g} use distant supervision to overcome domain gaps for identifying (non-)argumentative text.

\textbf{MTL} has been applied in many different
settings. \newcite{Bollmann2016} and \newcite{Peng2016} use data from different domains as different tasks and thereby improve historical spelling normalization and Chinese word segmentation and NER, respectively. \newcite{Plank2016} 
apply an MTL setup to POS tagging across 22 different languages, where the auxiliary task is to predict token frequency. 
\citet{Eger2017} explore sub-tasks (such as component identification) of a complex AM tagging problem (including relations between components) as auxiliaries and find that this improves performances. However, they stay within one single domain and dataset, and thus their approach does not address the question how new AM datasets with sparse data can profit from existing AM resources. 
Conceptually closest to our work, \newcite{Braud2017} leverage data from different languages as well as different domains in order to improve discourse parsing. 
While MTL 
was shown effective for syntactic 
tasks under certain conditions \cite{Sogaard2016}, 
\newcite{Alonso2017} find that MTL does not improve performances in four out of five semantic (i.e., higher level) tasks that they study. 
We are among the first to perform a structured investigation of MTL for higher-level pragmatic tasks, which are thought to be much more challenging than syntactic tasks \cite{Alonso2017}, and in particular, explore it for AM in cross-domain settings.

\section{Experiments}\label{sec:experiments}

 \begin{table*}[t]
\centering
\small
\begin{tabular}{p{3.4cm} p{3cm} r  r  
p{5.7cm}}
\toprule
\textbf{Dataset} & \textbf{Domain} & \textbf{\#Docs} & \textbf{Tokens} 
& \textbf{Component Types} \\
\midrule
\citet{Reed2008} & \texttt{var}ious ({Ar}aucaria) & 507 & 120 
& C (16), P (46), O (38)  \\ 
\citet{Biran2011} & \texttt{wiki}pedia discussions & 118 & 1592 
& C (9), justification (23), O (68) \\
 \newcite{LiuHEtAl2017},
 \newcite{GaoEtAl2017}
 & \texttt{hotel} reviews & 194 & 185 
 & 
 C (39), P (22), major C (7), implicit P (8),
 background (7), recommendation (5), O (12)
 \\
 \citet{Habernal2017a} & \texttt{web} discourse & 340 & 250 
 & C (4), P (25), backing (13), rebuttal (3), refutation (1), O (54) \\ 
 \citet{Habernal2017} & \texttt{news} {comm}ents & 1927 & 108 
 & P (53), O (47) \\
 \citet{Stab2017} & persuasive \texttt{essays} & 402 & 366 
 & C (15), P (45), major C (8), O (32) \\
 \bottomrule
\end{tabular} 
\caption{AM datasets: C -- claim, P -- premise, O -- non-argumentative; numbers in parentheses are label distributions in \%; `tokens' is the average in each document. 
}
\label{table:am_data}
\end{table*}

\paragraph{Data} We experiment with six datasets for \emph{argument component identification}, i.e. the token-level segmentation and typing of components.
These datasets are all of different sizes, have different average text lengths, and different argument component types and label distributions, as summarized in Table~\ref{table:am_data}. 
We only choose datasets containing both argumentative components and non-argumentative text.
Claims are available in five of six datasets, and 
all datasets have premises (resp. ``justification"), although it is unclear how large the conceptual overlap is across datasets. 
Further component types are idiosyncractic. \texttt{hotel} has the largest number of types, namely, six.
Most datasets also come with further information, e.g.\ relations between argument components, which are not considered here.

\paragraph{Approach} Due to the difference in annotations used in the different datasets, we consider each dataset as a separate AM task.
We treat all of them 
as sequence tagging problems,
where predicting BIO tags (argument segmentation) and argument component types (component classification) is framed as a joint task. This is achieved through token-level BIO tagging with the label set $\{O\}\cup\{B,I\}\times T$, where $T$ is a dataset specific set of argument component types, e.g. $T=\{\textrm{claim},\textrm{premise},\ldots\}$.
Thus, the overall number of tags in each dataset is twice the number of non-``O'' component types plus one ($2\cdot|T|+1$).
We use the state-of-the-art framework by \citet{Reimers2017a} for both single-task learning (STL) and MTL.
It employs a bidirectional LSTM (BILSTM) model with a CRF layer over individual LSTM outputs to account for label dependencies. 
We use \emph{nadam} as optimizer. 
For MTL, the recurrent layers of the deep BILSTM are shared by all tasks, with a separate CRF layer for each task. All tasks terminate at the same level.
The main task determines the number of mini-batches used for training, i.e.\ in every iteration the main task is trained on all its mini-batches and all other (auxiliary) tasks are trained on the same number of (randomly drawn) mini-batches.

To simulate data sparsity, we experiment with different sizes of training data for the main task.
We first draw a ``sparse'' training set of 21K tokens\footnote{Or more, since whole documents are added to the training set until the sum of tokens is at least 21K.
Similarly for smaller training and dev sets.} for each of the six AM datasets and a dev set of 9K to simulate a sparse scenario with 30K given tokens. 
The remaining data of each specific dataset is used as its test set (at least 5K tokens).
We then randomly draw a subset of the training data to create three more `sparsity scenarios' with 12K, 6K, and 1K tokens, respectively. 
Both dev and test set remain the same as in the 21K scenario. 
It is worth emphasizing how little data is used in the 1K scenario---only 1-10 documents (or roughly 50 sentences).
We train a separate STL system for each of the six datasets and each of the four sparsity scenarios.
In the MTL setup, the respective sparsity data is used as the main task, all other (auxiliary) AM datasets, each considered a separate task, are trained on all their available data. 
To measure the effect of MTL as opposed to a mere increase of training data, we furthermore train for each main task (i.e.\  each dataset and sparsity scenario) an STL system on the union of (training data of) main and auxiliary task, and evaluate it on the main task's test data.

\paragraph{Hyperparameter optimization} For each sparsity scenario and dataset we train 50 STL/MTL systems using GloVe embeddings \cite{Pennington2014} and 50 using the embeddings by \citet{Komninos2016}.
For each run we randomly choose a layout with either one hidden layer of $h\in\{50,100,150\}$ units or two layers of 100 units as well as
variational dropout rates  between 
0.2 and 0.5 for the input layer and for the hidden units.

\section{Results}\label{sec:results}
Note that we experiment with artificially shrunk datasets, which makes our results incomparable with those reported for the full datasets in other works. Nevertheless, it is to be expected that our STL model is on par with results obtained in recent works also using neural models for argument component identification, since our state-of-the-art BILSTM has the same architecture as the one by \citet{Eger2017}.

\paragraph{Overall trends} Table~\ref{tab:results-am} reports the 
average macro-F1\footnote{As implemented in scikit-learn \cite{scikit-learn}.} test scores over the respective ten best (according to the macro-F1 dev scores) hyperparameter configurations.
We compare 
STL on each task, MTL with all remaining datasets as auxiliary tasks, and the 
union baseline.
For three of the six datasets, MTL yields a significant improvement in all sparsity scenarios. Interestingly, these are the datasets with only one or two types of argument components. 
For the other three datasets, MTL only yields an improvement in the sparser data scenarios.
The union baseline generally performs (considerably) worse than STL in all scenarios. 
This implies that the domains and component types (label spaces) used in the different AM datasets are too diverse to 
model them as one single task
and that the improvement of MTL over STL cannot be attributed to more available data.

\begin{table}[t!]
\small
\begin{tabular}{lllll}
\toprule
Dataset & 21K & 12K & 6K & 1K \\
\midrule
\texttt{var} -- STL & 43.34 & 42.85 & 38.89 & 31.30 \\
\texttt{var}  -- MTL & \textbf{47.39} & \textbf{45.63} & \textbf{42.14} & \textbf{37.10} \\
\texttt{var}  -- BL & 30.45 & 27.35 & 26.75 & 26.62\\
\midrule
\texttt{wiki} -- STL & 23.37 & 22.57 & 20.93 & 19.74 \\
\texttt{wiki}  -- MTL & \textbf{32.50} & \textbf{31.99} & \textbf{28.03} & \textbf{20.17\textsuperscript{*} }\\
\texttt{wiki}  --BL & 18.34 & 18.12 & 17.49 & 20.47\\
\midrule
\texttt{news}  -- STL & 56.49 & 54.61 & 54.21 & 49.67 \\
\texttt{news} -- MTL & \textbf{57.76} & \textbf{56.34} & \textbf{55.41\textsuperscript{*}} & \textbf{52.43} \\
\texttt{news} -- BL & 32.63 & 40.63 & 36.54 & 35.51 \\
\midrule
\texttt{essays} -- STL & 60.54 & 56.35 & 49.68 & 24.60 \\
\texttt{essays} -- MTL & 60.55 & \textbf{57.90\textsuperscript{*}} & \textbf{52.14} & \textbf{32.55} \\
\texttt{essays} -- BL & 48.38 & 31.58 & 21.13 & 12.39 \\
\midrule
\texttt{web} -- STL & 23.43 & 22.33 & 19.71 & 11.28\\
\texttt{web} -- MTL & 23.27 & 22.97 & \textbf{21.73} & \textbf{15.31}\\
\texttt{web} -- BL & 15.21 & 14.94 & 12.09 & 10.80 \\
\midrule
\texttt{hotel} -- STL & 47.91 & 47.78 & 45.64 & 29.82\\
\texttt{hotel} -- MTL & 46.44 & 46.78 & 46.60 & \textbf{39.45} \\
\texttt{hotel} -- BL & 45.69 & 43.61 & 42.56 & 20.39\\
\bottomrule
\end{tabular} 
\caption{Macro-F1 for AM component identification,
comparing MTL, STL (significant differences in bold with $p < 0.0$1, $p < 0.05$ if \textsuperscript{*} using Mann-Whitney U Test) and union baseline (BL).
}
\label{tab:results-am}
\end{table}

\begin{figure}
  \input{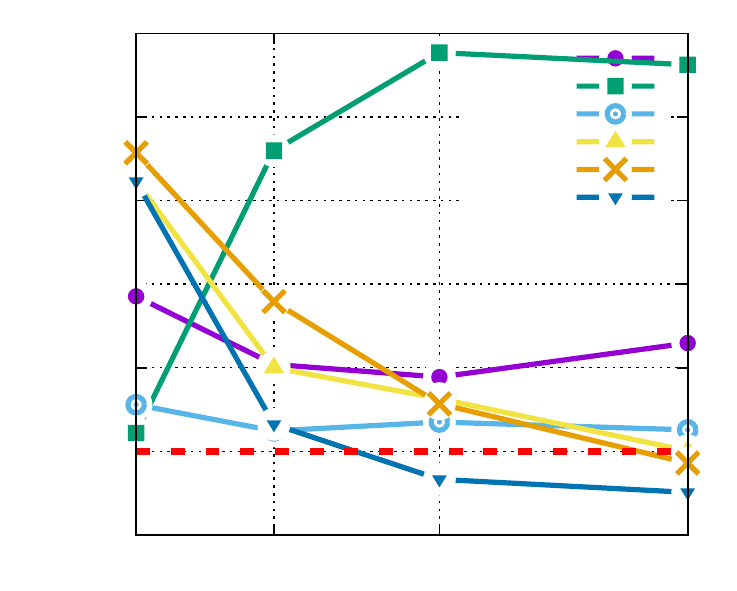}
  \caption{MTL versus STL: curves $\Delta(k)=\text{MTL}_{\text{norm}}(k)-\text{STL}_{\text{norm}}(k)$ as a function of size $k$ of main task.}
  \label{fig:cross-domain}
\end{figure}

Figure~\ref{fig:cross-domain} shows the general trends of our results. For each dataset, the figure plots the difference between normalized MTL and normalized STL macro-F1 scores ($\text{MTL}_{\text{norm}}(k)-\text{STL}_{\text{norm}}(k)$) 
for 
$k=$ 1K, 6K, 12K, 21K training data points for the main task.
For each specific dataset, the normalized macro-F1 score is defined as 
$\sigma_{\text{norm}}(k)=\frac{\sigma(k)}{\text{STL}(\text{1K})}$, where $\sigma(k)$ is the original macro-F1 score and STL(1K) denotes the STL score for 1K training data. Using this normalization, all scores are directly comparable and have the interpretation of improvement over the STL scenario with 1K tokens. 
It is noteworthy that MTL always improves over STL when the main task is very sparse (1K) and gains are sometimes substantial (between 30 and 40\% for \texttt{web}, \texttt{essays}, and \texttt{hotel}).

We observe three different patterns with respect to the main task: (i) for \texttt{essays}, \texttt{web}, and \texttt{hotel}, MTL is considerably better than STL when the main task is sparse, but 
for 21K tokens we observe either minimal gains or losses from MTL compared to STL.
(ii) The \texttt{var} and \texttt{news} datasets are stable, with consistent small gains from MTL over STL for all sizes of the main task. Finally, (iii) \texttt{wiki} displays an unusual pattern in that MTL gains are increasing with the amount of training data.
We attribute this to the large label imbalance in \texttt{wiki}, where nearly 70\% of the data is `O'.
When training data is very sparse, STL predicts 99\% of all tokens as `O', which results in a high F1 score for this component type but very low F1 scores (below 1\%) for the two other component types. The macro-F1 is thus lower than that of MTL, where `claim' and `premise' have a higher F1 score. Even though STL improves on the identification of `premise' and `claim' in the 21k scenario, the trend remains, since MTL also improves on these labels.

\paragraph{Detailed analysis} 
Upon closer inspection, we find that across all datasets MTL generally improves performance for class labels with low frequency as compared to STL. The more training data becomes available, the better STL gets in predicting such class labels, thus closing the gap to MTL. However, for \texttt{wiki} even 21K does not seem sufficient for STL to learn the two infrequent class labels, predicting 87\% as `O', so MTL still yields more than 10pp higher F1 for these infrequent classes.

Further analysis of our results 
reveals that the increase in the overall F1 score for MTL over STL is both due to improved component segmentation (BIO labeling) and better type prediction.
For example, in the 21K and 6K data settings, the BIO labeling improves by 1-4pp macro-F1 for nearly all datasets
and even by up to 17\%
for \texttt{wiki}. 
Unsurprisingly, in most cases, MTL also reduces invalid BIO sequences (`O' followed by `I'). 
Regarding the F1 scores of argument component types, we observe an improvement of MTL over STL for {claims} or {major claims} in all datasets containing these types and for {premises} in all but one dataset.
It is further interesting that
for the \texttt{hotel} dataset, MTL confuses premises mainly with the semantically similar implicit premises, whereas STL confuses premises with claims. Moreover, in both \texttt{hotel} and \texttt{essays}, claims are rarely predicted to be major claims, but major claims are predicted to be claims (both STL and MTL).

These results indicate that, despite the different domains and label spaces of the six datasets, MTL appears to learn generalized cross-domain representations of argument components, which aid argument component identification in sparse data scenarios and across domains.

\section{Concluding Remarks}\label{sec:conclusion}
We showed that MTL improves performance over STL on AM tasks
(particularly) when training data is sparse.
More precisely, argument component identification on a small AM dataset improves when treating other AM datasets as auxiliary tasks, even if these include different component types, coming from diverse domains. 
Overall, our results challenge the view that MTL is only infrequently effective for semantic or higher-level tasks \cite{Alonso2017}.
We attribute the success of MTL over STL to a few factors in our setting: (1) \citet{Alonso2017} used syntactic auxiliary tasks for semantic main tasks, whereas we choose only higher-level auxiliary tasks for higher-level main tasks. (2) The label spaces of all our tasks are relatively small, so that 
generalized representations can be learned in the 
LSTMs' hidden layers without suffering from \emph{label} sparsity. (3) The AM tasks considered here apparently do share common ground, a finding worth mentioning in itself given the contrary evidence in related work \cite{Daxenberger2017}.

Our findings cannot be 
readily anticipated by previous research, which has
reached mixed conclusions regarding the effectiveness of MTL overall and particular aspects, such as the size of main task. For example, while \citet{Luong2016} suggest that success of MTL requires that the auxiliary task does not swamp the main task data, \citet{Benton2017} and \citet{Yang:2017} come to the opposite conclusion that MTL is particularly effective when the data of the main task is small, and
\citet{Bingel2017} find a low correlation between size of the main task and MTL success.
Our curves in Figure \ref{fig:cross-domain} appear to \emph{prefer} the view that MTL is effective when the main task training data is sparse.

The scope for future work is vast. For example, 
it would be interesting to investigate whether standard low-level tasks, such as POS tagging or chunking,
are effective for AM.
Furthermore, other architectures for multi-task learning that apply soft parameter sharing, such as sluice networks \cite{Ruder2017}, will be investigated.

\section*{Acknowledgments}
Calculations for this research were conducted on the Lichtenberg high performance cluster of Technische Universit\"{a}t Darmstadt.
This work has been supported by the
German Federal Ministry of Education and Research
(BMBF) under the promotional references 16DHL1040 (FAMULUS),
01UG1816B (CEDIFOR), and 03VP02540 (ArgumenText).

\bibliography{library}
\bibliographystyle{style/acl_natbib}




\end{document}